\documentclass[10pt,twocolumn,letterpaper]{article}

\usepackage{cvpr}              

\usepackage{color}
\usepackage{booktabs}
\usepackage{multirow}
\usepackage{graphicx}
\usepackage{overpic}

\AtEndPreamble{
    \usepackage[capitalize]{cleveref}
}
\usepackage{wrapfig}

\definecolor{lightyellow}{rgb}{1,1, 0.8}
\definecolor{yellow}{rgb}{1,0.97, 0.65}
\definecolor{orange}{rgb}{0.9, 0.85, 0.7}
\definecolor{tablered}{rgb}{1, 0.7, 0.7}
\definecolor{turquoise}{cmyk}{0.65,0,0.1,0.3}
\definecolor{purple}{rgb}{0.65,0,0.65}
\definecolor{dark_green}{rgb}{0, 0.5, 0}
\definecolor{red}{rgb}{0.8, 0.2, 0.2}
\definecolor{darkred}{rgb}{0.6, 0.1, 0.05}
\definecolor{blueish}{rgb}{0.0, 0.3, .6}
\definecolor{light_gray}{rgb}{0.7, 0.7, .7}
\definecolor{pink}{rgb}{0.9, 0, 0.6}
\definecolor{greyblue}{rgb}{0.25, 0.25, 1}
\definecolor{teal}{rgb}{0.0, 0.4, 0.4}
\definecolor{chocolate}{rgb}{1.0, 0.4, 0.0}

\definecolor{1st}{rgb}{0.9, 0.65, 0.65}
\definecolor{2nd}{rgb}{1, 0.85, 0.85}
\definecolor{3rd}{rgb}{1.0, 0.95, 0.95}

\newcommand{\ws}[1]{{\color{pink}#1}}

\renewcommand{\ws}[1]{#1}

\newcommand{\images}{\mathbf{I}}
\newcommand{\image}{\mathbf{i}}
\newcommand{\inputimages}{\mathbf{P}}
\newcommand{\poses}{\mathbf{C}}
\newcommand{\pose}{\mathbf{c}}
\DeclareMathOperator*{\argmin}{arg\,min}

\newcommand{\loss}[1]{\mathcal{L}_\text{#1}}
\newcommand{\gsparam}{{\boldsymbol{\Phi}}}

\newcommand{\mask}{\mathbf{m}}
\newcommand{\masks}{\mathbf{M}}
\newcommand{\diffusion}{\mathcal{G}}
\newcommand{\scoring}{\mathcal{S}}
\newcommand{\rendering}{\mathcal{R}}

\newcommand{\hallucinationscore}{\mathbf{s}}
\newcommand{\multiviewencoder}{\mathcal{V}}
\newcommand{\multiviewfeature}{\mathbf{F}}
\newcommand{\augimages}{\tilde{\inputimages}}
\newcommand{\suppl}{\texttt{supplementary material}}

\renewcommand{\paragraph}[1]{\vspace{.4em}\noindent\textbf{#1}}

\usepackage[pagebackref,breaklinks,colorlinks,allcolors=cvprblue]{hyperref}
\usepackage{xcolor}
\usepackage{colortbl}
\usepackage{float}
\definecolor{cvprblue}{rgb}{0.21,0.49,0.74}

\def\confName{CVPR}
\def\confYear{2026}

\title{HAD: Hallucination-Aware Diffusion Priors for 3D Reconstruction
}

\author{
Xi Liu$^{1,2}$\footnotemark[1] \quad
Weiwei Sun$^{1}$\footnotemark[2] \quad
Zhou Ren$^{1}$  \quad
Chris Broaddus$^{1}$ \quad 
Siyu Huang$^{2}$\quad
Laurent Guigues$^{1}$ 
\\
\small
$^{1}$Amazon AWS \hspace{1pt}
$^{2}$Clemson University \hspace{1pt}
}

\begin{document}
\renewcommand{\thefootnote}{\fnsymbol{footnote}}

\twocolumn[{
\renewcommand\twocolumn[1][]{#1}%
\maketitle
\begin{center}
    \centering
    \vspace{-1.5em}
    \captionsetup{type=figure}
    \captionsetup[subfigure]{labelformat=empty}
    \addtocounter{figure}{-1}
    \begin{subfigure}[b]{0.99\textwidth}
         \centering
         \includegraphics[width=0.99\textwidth]{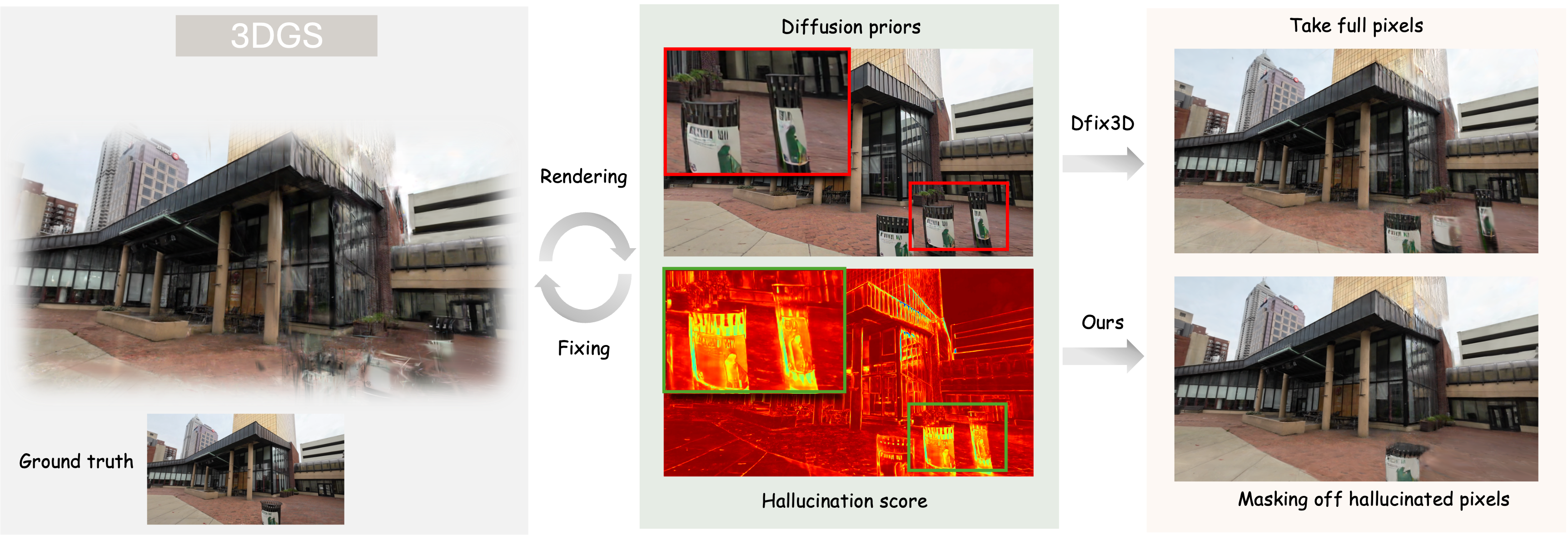}
    \end{subfigure}
          \hfill
    \captionof{figure}{ 
    While diffusion priors~\cite{wu2025difix3d+} enhance the quality of 3D reconstruction, they introduce detrimental \textit{aliens} -- the hallucinated elements that do not exist in the observed regions, as highlighted in boxes. This work addresses this issue through hallucination score modeling, achieving high-quality 3D reconstruction with improved fidelity.
    }
    \vspace{-0.5em}
    \label{fig:teaser}
\end{center}}]



\footnotetext[1]{Work done at Amazon.}
\footnotetext[2]{Corresponding author.}

\begin{abstract}
Diffusion priors have recently demonstrated strong capability in enhancing the quality of sparse-view 3D reconstruction by augmenting training views at novel viewpoints, 
but they inevitably introduce hallucinated content -- artifacts inconsistent with the input views -- into the final 3D model. 
To address this challenge, we propose Hallucination-Aware Diffusion prior (HAD), which estimates pixel-wise hallucination score maps for augmented images by leveraging multi-view reasoning capabilities from a feedforward novel view synthesis (NVS) network pre-trained on large-scale 3D data. These hallucination scores enable selective masking of unreliable pixels during the progressive 3D reconstruction procedure, preventing the introduction of non-existent artifacts into the 3D model. To further enhance performance, we create multiple versions of augmented images at each novel view by conditioning the diffusion prior on different input views, which are then fused into a final image that leverages the broader context across all input views. We show that our method substantially reduces hallucination artifacts in diffusion-assisted 3D reconstruction, thereby achieving state-of-the-art performance across multiple benchmarks on novel view synthesis. Our project are publicly available at \href{https://xiliu8006.github.io/HAD-Project-website/}{project website}.
\end{abstract}

\vspace{-1em}
\section{Introduction}
\label{sec:intro}
Neural Radiance Fields (NeRF)~\cite{mildenhall2020nerf} and 3D Gaussian Splatting (3DGS)~\cite{kerbl3Dgaussians} have emerged as groundbreaking approaches in 3D reconstruction, demonstrating remarkable capabilities in novel view synthesis (NVS) from multiple 2D views. Despite their significant achievements, these methods~\cite{mildenhall2020nerf,kerbl3Dgaussians,kheradmand20243d} require dense camera coverage and high-fidelity image inputs to maintain reliable performance. This dependency severely limits performance in data-sparse scenarios, such as sparse-view settings and extreme novel-view extrapolation tasks where the quality of rendered images degrades dramatically, as illustrated in \cref{fig:teaser}. One approach to address data sparsity is to leverage generative diffusion priors to augment novel-view data by removing artifacts from rendered images through denoising conditioned on the original input views~\cite{wu2025difix3d+,liu20243dgs,fischer2025flowr}.

While these augmented views achieve photorealism, they often fail to maintain fidelity to the original input views. The root cause is that the diffusion process, by design, does not strictly preserve the content of the conditional input views, resulting in multi-view inconsistency in the generated images. Consequently, when this hallucinated content from augmented views is incorporated into the 3D model, it leads to a {\it hallucination issue} in 3D reconstruction, where rendered views appear high-quality but exhibit low fidelity to the input views. While recent works ~\cite{fischer2025flowr,yin2025gsfixer, liu20243dgs} mitigate this issue by enhancing the ability of diffusion priors to respect input views -- e.g., through video diffusion models~\cite{zhou2025stable, wu2025genfusion}, multi-view inconsistency -- as shown in \cref{sec:hallucination—analysis_appendix} of \suppl -- still remains in generated images due to the generative nature of diffusion priors. 

To address this limitation, we propose incorporating hallucination awareness into the augmented views. Rather than preventing hallucination entirely, we retain the ability to filter out hallucinated content when incorporating the augmented images into 3D models. Accordingly, we coin our method as Hallucination-Aware Diffusion priors (HAD). Specifically, we propose a hallucination score network to predict pixel-wise hallucination scores for augmented views, quantifying the inconsistency between augmented views and the original input multi-views. The network consists of a multi-view encoder that understands multiple input views and subsequently a hallucination score branch that detects hallucinated content in augmented views.

Inspired by the ability of recent feedforward novel view synthesis (NVS) networks~\cite{jin2025lvsm} to reconstruct novel views with high fidelity, we leverage the pre-trained feature backbone of a well-trained NVS network as our multi-view encoder. The key insight is that these NVS feature backbones develop strong multi-view reasoning abilities through training on large-scale 3D data. By leveraging this pre-trained knowledge, we can train our hallucination network on a small-scale curated dataset, bypassing the challenging large-scale data curation that would otherwise be required—each training sample must include both input multi-views and augmented novel views, necessitating a complete pipeline of 3D reconstruction, novel view rendering, and diffusion prior denoising.

Our network identifies hallucinated content in augmented images through the hallucination score map, enabling the masking of hallucinated pixels in the augmented views during 3D model optimization. To further enhance the performance of HAD, we propose a multi-sampling strategy that generates multiple images at the same novel view by conditioning the diffusion process on different input views, thereby incorporating a broader context from the input views into HAD. We then fuse these images into an improved image for 3D reconstruction using the hallucination score maps. This strategy increases the number of conditional input views in the diffusion prior without retraining, effectively reducing the ratio of hallucinated content in diffusion-generated images.

We validate our method on two standard NVS benchmarks: DL3DV~\cite{ling2024dl3dv}, on which we train HAD for in-domain evaluation, and MipNeRF 360~\cite{barron2022mip} for cross-domain evaluation. 
As demonstrated in~\cref{sec:experiments}, our method significantly outperforms state-of-the-art approaches, with PSNR improvements of 0.78dB on DL3DV and 0.69dB on MipNeRF 360. We then summarize our contributions as below:
\begin{itemize}
\item We identify a critical limitation where diffusion priors, while alleviating data sparsity in 3D reconstruction, introduce hallucination issues that compromise fidelity to input views despite achieving photorealistic rendering. 
\item We propose Hallucination-Aware Diffusion priors (HAD), the first framework to explicitly tackle the hallucination issue in diffusion-assisted 3D reconstruction through hallucination score modeling. To the best of our knowledge, this is the first work to study hallucination score modeling in this context.
\item We introduce a multi-sampling strategy into HAD that generates and fuses multiple versions of a novel view into a single improved image.
\item Our method achieves state-of-the-art performance in both in-domain and cross-domain evaluations. Extensive experiments show that our improvements stem from effective hallucination mitigation across diverse scenarios.
\end{itemize}

\section{Related Works}
\vspace{-0.5em}
\label{sec:related_work}
\paragraph{Radiance fields for novel view synthesis.}
Novel view synthesis (NVS) aims to render photorealistic images from novel viewpoints given a set of posed input images. Among recent approaches, Neural Radiance Fields (NeRF)\cite{mildenhall2020nerf} and 3D Gaussian Splatting (3DGS)\cite{kerbl3Dgaussians} have emerged as the two dominant paradigms. Subsequent research~\cite{barron2021mip,barron2022mip, chen2022tensorf, fridovich2022plenoxels, yu2021plenoctrees, mubarik2023hardware} has focused on addressing NeRF's limitations in handling unbounded scenes and improving its rendering efficiency through hierarchical sampling, spatial factorization, and hybrid representations. 3DGS~\cite{kerbl3Dgaussians} leverages an efficient rasterization-based differentiable splatting pipeline~\cite{10.1145/383259.383300}, enabling real-time novel view synthesis with competitive visual quality. Following its success, numerous works~\cite{Huang2DGS2024, wu2025sparse2dgs, chen2024pgsr, yu2024mip, fan2024lightgaussian} have further enhanced 3DGS by improving depth and surface estimation accuracy, reducing storage overhead, and alleviating aliasing artifacts, collectively establishing it as one of the most powerful and practical frameworks for 3D scene reconstruction. In this work, we validate the effectiveness of our hallucination-aware diffusion priors on enhancing 3DGS. 

\paragraph{Novel view synthesis with priors.}
Recent advances demonstrate that incorporating priors from large foundation models can greatly enhance 3D reconstruction under sparse or imperfect observations. These priors provide strong guidance on geometry, texture, and semantics, enabling 3D reconstruction to move beyond reliance on purely original input data. Geometry-based methods~\cite{li2024dngaussian,wang2023sparsenerf,niemeyer2022regnerf,long2022sparseneus, yang2023freenerf, Jain_2021_ICCV} impose additional constraints such as depth, frequency, or semantic consistency to stabilize training, but their performance is often limited by inaccurate or inconsistent estimations. Image diffusion-based approaches~\cite{poole2023dreamfusion,yi2023gaussiandreamer, liu2023zero, shi2023MVDream, wu2025difix3d+} leverage pretrained diffusion models as powerful generative priors to reconstruct 3D scenes from text or very limited input views.
More recently, video and multi-view diffusion models~\cite{wu2024reconfusion,melaskyriazi2024im3d,blattmann2023stable, liu20243dgs,liu2026reconx, gao2024cat3d,chen2024mvsplat} further enhance view and temporal consistency through camera-conditioned generation. Nevertheless, while these diffusion priors effectively augment 3D reconstruction, they also introduce hallucinated content. We propose injecting hallucination awareness into diffusion-generated images and demonstrate that it significantly improves reconstruction performance.

\paragraph{Feedforward novel view synthesis.} 
With the remarkable success of Transformer architectures~\cite{vaswani2017attention} in both large language models and vision foundation models~\cite{dosovitskiy2020image}, feedforward networks~\cite{jiang2025anysplat,xu2025depthsplat,jin2025lvsm,jiang2025rayzer} have recently shown strong potential in novel view synthesis. The recent state-of-the-art LVSM~\cite{jin2025lvsm} demonstrates that a pure transformer-based network can synthesize high-quality images from sparse inputs, while Rayzer~\cite{jiang2025rayzer} further eliminates the need for camera poses during training while still achieving competitive or superior performance. Due to their reconstruction nature, NVS networks are trained to synthesize novel view images with high fidelity to the input views and thus are prone to high-frequency information loss, whereas diffusion priors, despite their low fidelity to input data, tend to generate high-frequency information. Thus, in this work, we leverage the feature backbone of a well-trained NVS network as our multi-view encoder for understanding the context of input views to predict reliable hallucination scores for diffusion-generated images. This approach operates under the mild assumption that NVS networks possess the ability to effectively understand input context in order to render novel views with high fidelity.

\ws{\paragraph{Uncertainty estimation in 3D reconstruction.} 
Recent works~\cite{shen2024estimating, goli2024bayes,amini2024instant,guo2024uc,NEURIPS2024_a076d0d1} estimate the uncertainty in 3D reconstruction, a concept related to our hallucination score.  However, unlike these methods that quantify uncertainty in the final 3D representation, our method directly suppresses hallucinations in the augmented novel views for reconstruction, thereby reducing the 3D uncertainty propagation into the 3D model.        
}
\begin{figure*}
\vspace{-1.0em}
  \centering
  \includegraphics[width=\linewidth]{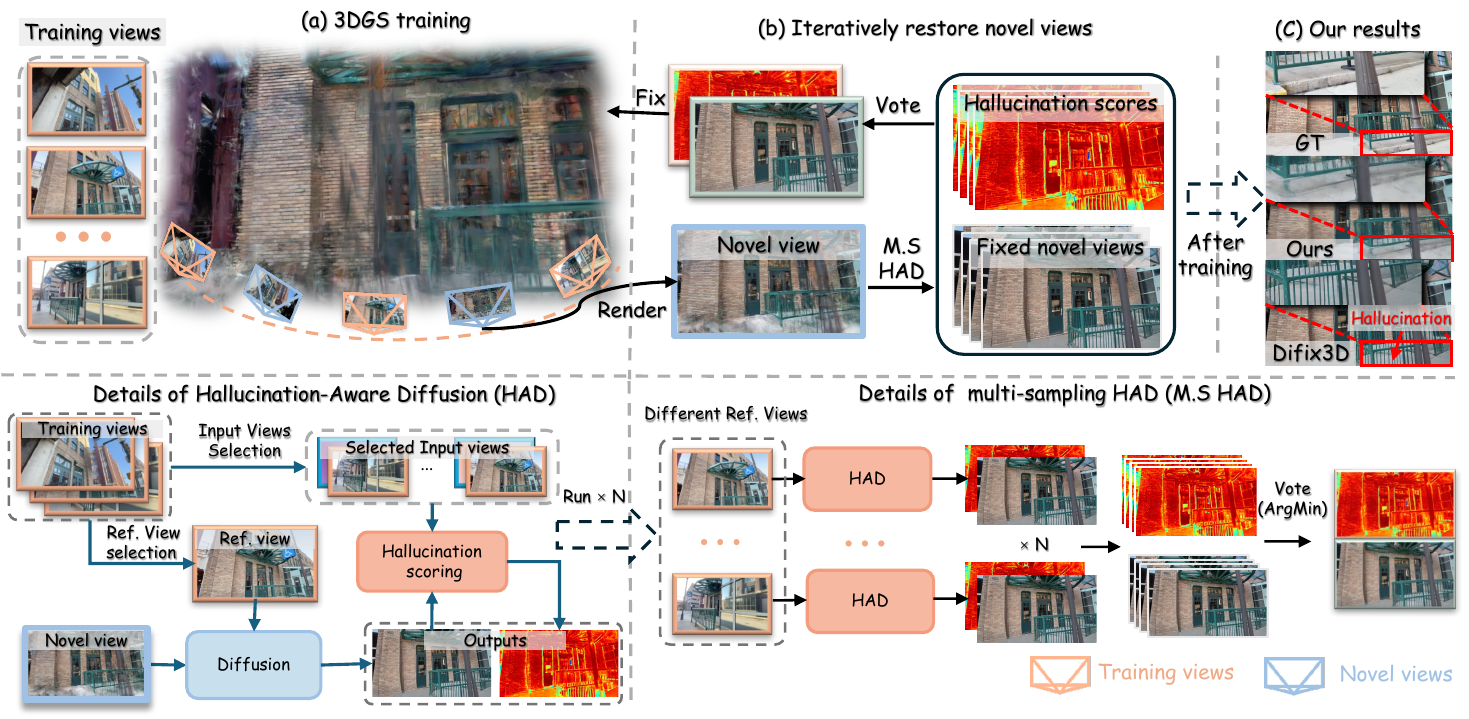}
\caption{
{\bf Overview of framework -- } 
We train 3DGS with input images and HAD-augmented novel views. HAD combines a pretrained diffusion prior (which generates images from 3DGS-rendered views conditioned on reference input images) with our hallucination score network (which predicts pixel-wise reliability maps). Our multi-sampling strategy fuses multiple generated versions into refined augmented views. Hallucination scores guide 3DGS optimization by masking off unreliable content, thus improving reconstruction quality.
}
\label{fig:framework}
\vspace{-1.4em}
\end{figure*}


\section{Preliminary}
\label{sec:preliminary}
We briefly describe the preliminaries for 3D Gaussian Splatting (3DGS) -- the 3D pipeline that we use to validate our HAD, feedforward novel view synthesis network and diffusion priors.

\paragraph{3D Gaussian Splatting.} 
Given a set of calibrated images with known camera parameters, the goal of 3D Gaussian Splatting (3DGS)~\citep{kerbl3Dgaussians} is to render photorealistic image from arbitrary viewpoints in real time. 3DGS models the scene as a collection of anisotropic Gaussian primitives $\{ (\boldsymbol{\mu}_i, \mathbf{S}_i, \mathbf{R}_i, \delta_i, \mathbf{c}_i) \}_{i=1}^{N}$ where $\boldsymbol{\mu}_i \in \mathbb{R}^3$ denotes the 3D center of the $i$-th Gaussian, $\mathbf{S}_i = \mathrm{diag}(s_x, s_y, s_z)$ is a scaling matrix, and $\mathbf{R}_i$ is a rotation matrix derived from a quaternion parameterization. Each Gaussian also carries an learnable opacity scale $\delta_i$ and a set of spherical harmonics (SH) coefficients $\mathbf{c}_i$ for modeling view-dependent color. The opacity $\alpha_i$ is expressed as
\begin{equation}
    \alpha_i = \delta_i \exp\!\left[-\tfrac{1}{2}(\mathbf{x}-\boldsymbol{\mu}_i)^{\!\top}\!\boldsymbol{\Sigma}_i^{-1}(\mathbf{x}-\boldsymbol{\mu}_i)\right],
\end{equation}
where the covariance matrix is given by $\boldsymbol{\Sigma}_i = \mathbf{R}_i \mathbf{S}_i \mathbf{S}_i^{T} \mathbf{R}_i^{\!\top}$.

During rendering, tile-based rasterization is used to identify, for each pixel, which Gaussians contribute to its color. And then each pixel  $\mathbf{p}$ can be rendered by blending N ordered points overlapping the pixel:
\vspace{-0.3em}
\begin{equation}
C(\mathbf{p}) = \sum_{i=1}^{N} \mathbf{c}_i\, \alpha_i \prod_{j=1}^{i-1} (1-\alpha_j)
\label{eq:alpha_composition}
\end{equation}

\vspace{-0.3em}
\paragraph{Feed-forward novel view synthesis network.}
A feedforward NVS network is a generalizable network that takes multiple views as input and outputs a 3D feature, enabling the rendering of images from novel viewpoints. We formulate it as:
\begin{equation}
    \image_{\pose} = \text{FFD}(\inputimages \mid \pose),
\end{equation}
where $\inputimages$ denotes the input calibrated images, $\pose$ represents novel view pose and $\image_{\pose}$ is the rendered image from $\pose$. Recent works~\cite{jiang2025anysplat,xu2025depthsplat} adopt a \emph{Gaussian head} to predict the parameters of a Gaussian set and then render the target image using~\cref{eq:alpha_composition}. In contrast, other approaches such as LVSM~\cite{jin2025lvsm}-- the state-of-the-art NVS network -- directly regresses the target image without explicitly reconstructing the intermediate 3D representation, leading to a more simplified and scalable framework than the explicit Gaussian head approach. In light of this, our hallucination score network leverages the pretrained feature backbone of LVSM~\cite{jin2025lvsm} to extract 3D information from input views.

\paragraph{Diffusion priors for 3D reconstruction.}
Recent advances have demonstrated that diffusion-based priors are highly effective for improving 3D reconstruction and scene enhancement~\cite{liu2026reconx, liu20243dgs, wu2025difix3d+, chen2024mvsplat360, fischer2025flowr}. By learning to model the data distribution $p_{\text{data}}(x)$ through an iterative denoising process, diffusion models can transform imperfect novel-view images into samples that follow the distribution of realistic images, under the assumption that the distribution of artifacts in imperfect novel-view images rendered by 3DGS is close to the noise distribution~\cite{wu2025difix3d+}. This property allows diffusion models, with light fine-tuning, to effectively remove artifacts in the images rendered by 3DGS at novel views.

Formally, a diffusion model progressively corrupts clean data $x \sim p_{\text{data}}$ with Gaussian noise to obtain noisy samples 
\begin{equation}
    x_\tau = \alpha_\tau x + \sigma_\tau \epsilon, \quad \epsilon \sim \mathcal{N}(0, I),
\end{equation}
where $\tau$ denotes the diffusion timestep, and $(\alpha_\tau, \sigma_\tau)$ control the noise schedule. 
The denoising network $\epsilon_\theta$ learns to invert the noising process by predicting the noise term $\epsilon$ added at each step, optimized via the denoising score matching objective:
\begin{equation}
    \mathbb{E}_{x \sim p_{\text{data}},\, \tau \sim p_\tau,\, \epsilon \sim \mathcal{N}(0,I)} 
    \Big[ \| \epsilon - \epsilon_\theta(x_\tau; c, \tau) \|_2^2 \Big],
\end{equation}
where $c$ denotes conditioning views sampled from the original input multi-views. The diffusion time $\tau$ is typically discretized into $T$ steps (e.g., $T=200$) with $\tau \sim \mathcal{U}(0, T)$, ensuring that the data becomes nearly Gaussian at the final step. In this work, we validate our hallucination awareness approach on Difix3D~\cite{wu2025difix3d+}, a single-step diffusion prior that achieves the top performance on NVS from sparse views.
\section{Methodology}
\label{sec:method}
Given $N$ calibrated images of a scene $\inputimages = \{(\images_{n}, \poses_{n}) | {n \in \{0, \ldots N \}}\}$, where $\images$ and $\poses$ denote the RGB images and camera poses respectively, 
our goal is to reconstruct a high-fidelity 3DGS model capable of producing high-quality renderings for both training views and ill-constrained novel views with insufficient observations.
Specifically, we train the 3DGS model under supervision from both the input views and augmented views generated by our hallucination-aware diffusion prior (HAD).

\subsection{3DGS training}
We formulate the 3DGS training as 
\begin{equation}
\argmin_{\gsparam} \lambda_{\text{input}} \loss{input} + \lambda_{\text{novel}} \loss{novel}  
\end{equation}
where $\loss{input}$ and $\loss{novel}$ are the rendering losses for input views and augmented novel views, respectively, $\lambda_{\text{input}}$ and $\lambda_{\text{novel}}$ are the coefficients for the two loss terms, and $\gsparam$ denotes the parameters of the 3DGS model described in \cref{sec:preliminary}. 

We follow 3DGS~\cite{kerbl3Dgaussians} to compute the rendering loss as in  at input views by combining $\loss{1}$ and $\loss{D-SSIM}$:
\begin{equation}
   \loss{input} = 0.8 \loss{1}\left(\rendering_{\gsparam}\left(\pose \right),  \image \right) + 0.2 \loss{D-SSIM}\left(\rendering_{\gsparam}\left(\pose\right),  \image\right)
\end{equation} 
where $\image$ and $\pose$ are the image and camera pose sampled from the input views, and $\rendering_{\gsparam}$ is the rendering function parameterized by $\gsparam$ that takes camera pose $\pose$ as input.

Our hallucination-aware diffusion prior (HAD) augments novel views to form $\loss{novel}$. Specifically, we follow Difix3D~\cite{wu2025difix3d+} to sample novel view poses by progressively departing from the input poses toward the target poses. HAD then generates the images and their corresponding hallucination score maps. We denote the resultant hallucination masks, poses, and images of $T$ novel views as $\augimages = \{(\masks_{t}, \tilde{\images}_{t}, \tilde{\poses}_{t}) \mid t \in \{0, \ldots, T\}\}$. We then calculate $\loss{novel}$ as: 
\begin{equation}
\begin{split}
    \loss{novel} =  &\loss{1}(\lnot \mask \odot  \rendering_{\gsparam}\left(\tilde{\pose} \right) , \lnot \mask \odot \tilde{\image}) \\
    &+ \loss{D-SSIM}(\lnot \mask \odot  \rendering_{\gsparam}\left(\tilde{\pose} \right) , \lnot \mask \odot \tilde{\image})
\end{split}
\end{equation}
where $\odot$ denotes the Hadamard product, $\lnot$ inverts the binary mask, and $\mask$, $\tilde{\pose}$, and $\tilde{\image}$ are the hallucination mask, pose, and RGB image, respectively, of a novel view generated through HAD, which we detail in \cref{sec:had}.

We illustrate the training process in \cref{fig:framework}. Specifically, unlike Difix3D~\cite{wu2025difix3d+}, which employs a two-phase training strategy that first fully trains a 3DGS model and then progressively updates it with diffusion priors using a small learning rate, we adopt a simplified single-phase training approach. Our method trains the 3DGS model with both input views and diffusion-augmented views simultaneously from the start. The key insight is that our hallucination score maps enable masking of unreliable pixels, allowing us to directly incorporate diffusion-augmented views from the beginning of training without requiring a pre-trained 3DGS initialization.

Nevertheless, we follow Difix3D~\cite{wu2025difix3d+} in alternating between view augmentation and training steps. In the view augmentation step, we sample novel view camera poses by interpolating between input and target views. 
HAD then generates augmented images $\tilde{\image}$ along with their corresponding pixel-wise hallucination score maps $\hallucinationscore$, which are converted into binary masks $\mask$ via simple thresholding.

\subsection{Hallucination-Aware Diffusion Prior}
\label{sec:had}
To enhance novel view synthesis quality, we propose the hallucination-aware diffusion prior (HAD) to augment images rendered at novel views and optimize the 3DGS model with these augmented views. Specifically, HAD consists of the diffusion prior (\cref{sec:diffusionprior}) from Difix3D~\cite{wu2025difix3d+} and our novel hallucination score estimator (\cref{sec:hall_score_net}). As shown in \cref{fig:framework}, the diffusion prior enhances novel views by removing artifacts through a denoising step conditioned on a sampled input view close to the novel view, while the hallucination score network predicts hallucination masks for the augmented views. Furthermore, to improve performance, we propose a multi-sampling strategy (\cref{sec:multisampling}) that creates multiple versions of augmented views by conditioning the diffusion prior on different input views, then fuses them into a higher-quality image.

\subsubsection{Refining Novel Views w/ Diffusion Prior}
\label{sec:diffusionprior}
Formally, given the novel view camera $\tilde{\pose}$, the pretrained diffusion prior generates the novel view image $\tilde{\image}_{\diffusion}$ as
\begin{equation}
\tilde{\image}_{\diffusion} = \diffusion\left(\rendering_{\Phi}\left(\tilde{\pose}\right)\right | \image_{\text{ref}})
\label{eq:render}
\end{equation}
where $\image_{\text{ref}}$ is a reference view sampled from the original input multi-views serving as conditioning 3D context for the view enhancement. The diffusion generated $\tilde{\image}_{\diffusion}$, while photorealistic, tend to introduce aliens that do not exist in the input multi-views.   
\subsubsection{Hallucination Score Estimation}
\label{sec:hall_score_net}

Thus, we propose a hallucination score network to estimate a pixel-wise hallucination score map $\hallucinationscore$ for the generated image $\tilde{\image}_{\diffusion}$, enabling hallucination-aware diffusion priors for 3DGS enhancement. Specifically, the hallucination score network consists of two components: a multi-view feature encoder $\multiviewencoder$ that processes multiple input views, and a score estimation branch $\scoring$ that predicts hallucination scores for novel view images using both the multi-view features and the novel view images. \ws{For further details, please refer to \cref{sec:model_arch_appendix} of \suppl.}

To leverage the multi-view reasoning capability of existing novel view synthesis networks, we base our hallucination score network on the pre-trained LVSM~\citep{jin2025lvsm}, a state-of-the-art feedforward NVS model with a feature backbone that processes multiple input views and an NVS head for image generation. 
We set the pretrained feature backbone of LVSM~\cite{jin2025lvsm} as $\multiviewencoder$ and realize the $\scoring$ as a simple three layer UNet. 
We train the framework with the $\multiviewencoder$ frozen on a small curated dataset of novel view and original multi-view pairs.

Formally, we calculate the hallucination score as
\begin{equation}
    \hallucinationscore = \scoring_\theta ( \tilde{\image}_{\diffusion} | \rendering_{\gsparam} (\tilde{\pose}), \multiviewfeature_{\tilde{\pose}} )
\end{equation}
where $\theta$ denotes the parameters of $\scoring$, $\tilde{\pose}$ is the novel view pose, and $\multiviewfeature_{\tilde{\pose}}$ represents the multi-view features, which we formulate as:
\begin{equation}
   \multiviewfeature_{\tilde{\pose}} = \multiviewencoder(\inputimages \mid \tilde{\pose})
\end{equation}
where $\inputimages$ denotes the original multiple input views. Thus, the multi-view encoder $\multiviewencoder$ outputs features aggregated at the novel view pose $\tilde{\pose}$ from the input views.

We then concatenate $\multiviewfeature_{\tilde{\pose}}$, $\tilde{\image}_{\diffusion}$, and optionally the 3DGS-rendered image $\rendering_{\gsparam}(\tilde{\pose})$, from which $\scoring$ predicts a pixel-wise hallucination score map $\hallucinationscore$ that reflects the hallucination probability of each pixel in the diffusion-augmented image $\tilde{\image}_{\diffusion}$.
To train $\scoring$, we define the ground-truth hallucination score as the mean absolute error (MAE) between the diffusion-generated image and the ground-truth image. We supervise $\scoring$ with an L2 loss between the predicted score and the ground-truth score.

\subsubsection{Multi-Sampling Strategy}
\label{sec:multisampling}
To further enhance HAD, we propose a multi-sampling strategy that creates multiple versions of augmented views and fuses them to produce higher-quality novel views for 3DGS optimization. Specifically, the diffusion model augments novel views with different sampled conditioning input views, generating multiple augmented views along with their corresponding hallucination score maps.

In Difix3D~\citep{wu2025difix3d+}, each diffusion refinement relies solely on the nearest reference image $\image_{\mathrm{ref}}$, which limits the exploitation of complementary cues from other viewpoints. Our multi-sampling strategy creates multiple versions of augmented images—$\{(\tilde{\image}_{\diffusion}^k, \hallucinationscore^k) \mid k \in \{1, \ldots, K\}\}$—by conditioning on $K$ sampled references from the input views. We then fuse the images by selecting pixels with the lowest hallucination score across all candidate versions: 
\[ \tilde{\image}[i] = \tilde{\image}_{\diffusion}^{k^\ast}[i], \quad k^\ast = \arg\min_k \hallucinationscore^k[i], \] 
where $[i]$ denotes pixel indexing, $\tilde{\image}_{\diffusion}^k$ and $\hallucinationscore^k$ denote the diffusion-generated image and its predicted hallucination score from the $k$-th reference view. This pixel-wise selection effectively fuses reliable regions from multiple refinements, yielding augmented novel views with improved fidelity. More importantly, this strategy allows the diffusion prior to exploit broader multi-view information, effectively reducing hallucination issues in the final 3DGS model.

\section{Experiments}
\newcommand{\dldv}{\texttt{DL3DV}~\cite{ling2024dl3dv}}
\newcommand{\mipnerf}{\texttt{MipNeRF360}~\cite{barron2022mip}}

\label{sec:experiments}
\subsection{Experimental Setup}
\paragraph{Dataset and metrics.} 
We primarily use Peak Signal-to-Noise Ratio (PSNR), structural (SSIM~\cite{wang2004image}) and perceptual (LPIPS~\cite{zhang2018unreasonable}) similarities as metrics to quantify the performance of novel view synthesis. 
And we conduct experiments on two well-known datasets. 
\begin{itemize}
    \item \dldv~is a large-scale scene dataset containing calibrated multi-view images. We first curate training dataset of randomly selected 116 scenes from benchmark dataset for hallucination score network training. To validate the performance of HAD on novel view synthesis, we follow Difix3D~\cite{wu2025difix3d+} to select other 24 scenes for testing. For each testing scene, we partition the available views into three subsets: training, target, and test views. Specifically, we select nine sparse views as input for 3D reconstruction. The remaining views are then divided into test and target subsets. We follow ~\cite{wu2024reconfusion} sample test views randomly from the unused views at a ratio of 25\%, and compute the final performance metrics, providing a representative assessment of novel-view reconstruction quality. 
    Target views are uniformly sampled from the last few viewpoints of each scene at a ratio of 50\%. The target views are used to sample novel views for generative prior augmentation in 3DGS training. This design ensures that no leaked information from the target subset is used for model adaptation, maintaining fair and unbiased comparisons across methods.
    \item \mipnerf~is a dataset comprising 9 scenes commonly used for evaluating novel view synthesis.
    We adopt the training and testing view splits provided by GenFusion~\cite{wu2025genfusion}, which consists of 9 training input views per scene. For the target views, similarly, we uniformly sample 50\% of the remaining unused views. 
\end{itemize}

\paragraph{Implementation details.} 
To construct the hallucination score network, we utilize the pretrained multiview encoder from the LVSM model~\cite{jin2025lvsm} and train an additional score prediction branch on a small curated dataset. 
We prepare the dataset using the Difix3D~\cite{wu2025difix3d+} pipeline. 
For each scene, we first train a 3DGS model using 9 input training views, then generate 100 augmented novel views via diffusion priors~\cite{wu2025difix3d+} at a resolution of 960×540. The poses for these augmented views are sampled from views excluded from the 3DGS training set, providing ground-truth images that enable computation of the ground-truth hallucination score map for each augmented view. We fine-tune the network for 10k iterations with a batch size of 2 per GPU, requiring approximately 28 hours on eight NVIDIA V100 32GB GPUs.  
For 3DGS training, we set the learning rates to $8e^{-5}$ for Gaussian means, $5e^{-2}$ for opacity, $1e^{-3}$ for rotation, $5e^{-4}$ for the 0-th order spherical harmonics (SH$_0$), and $2.5e^{-5}$ for higher-order spherical harmonics (SH$_N$), and train for 30k iterations. And we set $\lambda_{\text{input}}$ and $\lambda_{\text{novel}}$ as 1. \ws{We empirically set the threshold as 0.9 to calculate the $\mask$.}

\begin{table}
\centering
\caption{\textbf{Quantitative comparison of different methods on DL3DV \cite{ling2024dl3dv}.} Best, second, and third results are highlighted in \colorbox{1st}{1st}, \colorbox{2nd}{2nd}, and \colorbox{3rd}{3rd}, respectively. ($\uparrow$: higher is better, $\downarrow$: lower is better). 
Note that ours* denotes a variant following the two-phase 3DGS optimization strategy of Difix3D, enabling a fair comparison between diffusion priors with and without hallucination awareness. 
We denote "$+$" as post-rendering, which evaluates the rendering quality from 3DGS augmented by a diffusion model, rather than direct rendering from 3DGS.
}
\resizebox{\columnwidth}{!}{
\begin{tabular}{lccccc}
\toprule
 \textbf{Method} & \textbf{Type} & \textbf{PSNR}$\uparrow$ & \textbf{SSIM}$\uparrow$ & \textbf{LPIPS}$\downarrow$ \\
\midrule
 Depthsplat \cite{xu2025depthsplat} & FFD & \cellcolor{2nd} 18.324 & \cellcolor{2nd} 0.640 & \cellcolor{2nd} 0.378 \\
 LVSM \cite{jin2025lvsm} & FFD & \cellcolor{1st} 19.855 & \cellcolor{1st}0.636 & \cellcolor{1st}0.252 \\
\midrule
 Gsplat-3DGS \cite{kerbl3Dgaussians} & Optimization & 19.004 & 0.679 & 0.281 \\
 Gsplat-mcmc \cite{kheradmand20243d} & Optimization & {20.532} & {0.721} & {0.225} \\
 Difix3D \cite{wu2025difix3d+} & Optimization & \cellcolor{3rd}{21.355} & \cellcolor{3rd}{0.734} & \cellcolor{3rd}{0.199} \\
 Ours* & Optimization &  \cellcolor{2nd}21.983 & \cellcolor{2nd} 0.755 & \cellcolor{2nd} 0.195 \\
 Ours & Optimization & \cellcolor{1st}{22.134} & \cellcolor{1st}0.757 & \cellcolor{1st}0.190 \\
 \midrule
 Difix3D+ \cite{wu2025difix3d+} & Post-rendering & \cellcolor{2nd}20.673 & \cellcolor{2nd}0.709 & \cellcolor{2nd} 0.181 \\
 Ours+ & Post-rendering & \cellcolor{1st}21.213 & \cellcolor{1st}0.719 & \cellcolor{1st} 0.177 \\
\bottomrule
\end{tabular}
}
\vspace{-1.2em}
\label{tab:main_result}
\end{table}

\begin{figure*}
\vspace{-0.5em}
  \centering
  \includegraphics[width=\linewidth]{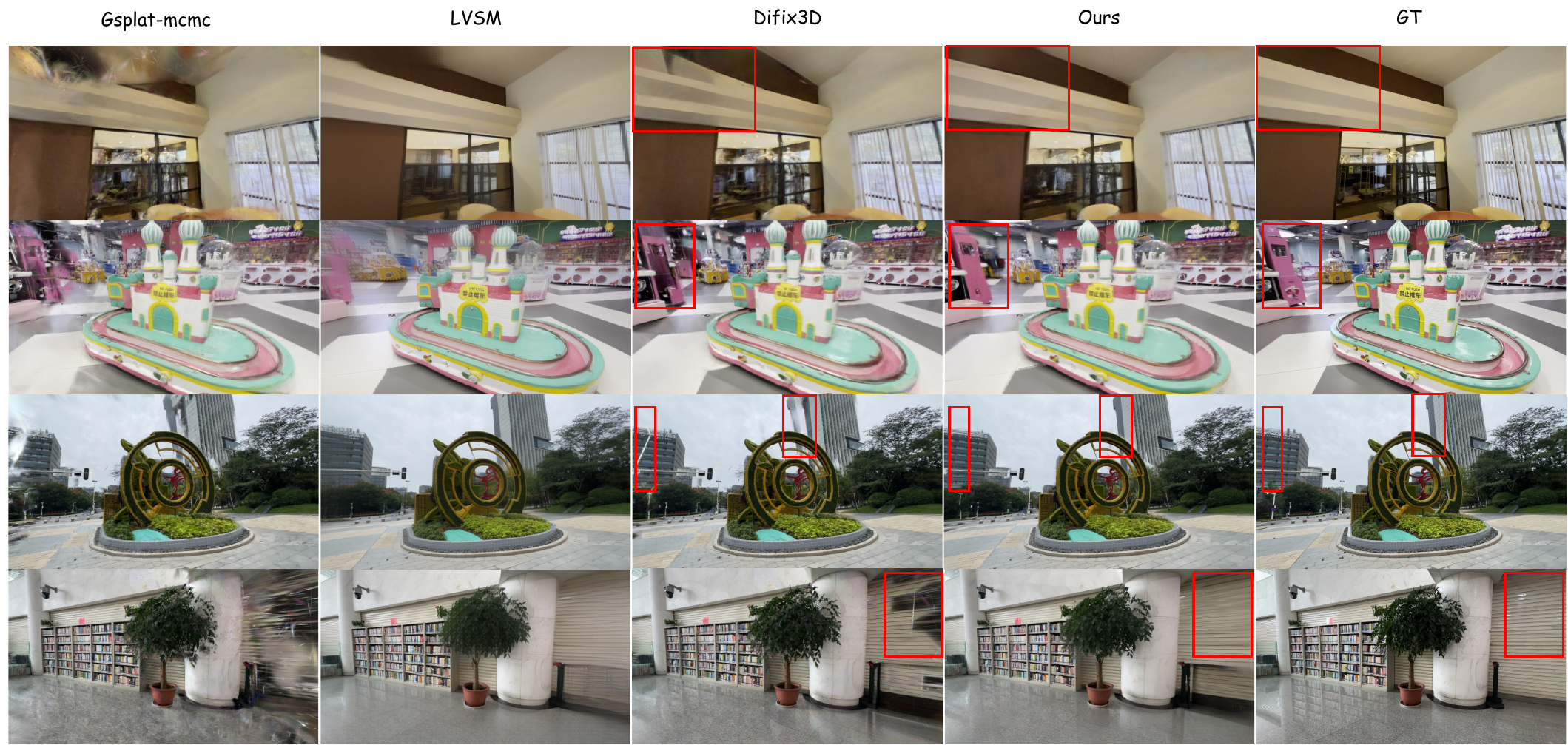}
\caption{
\textbf{Examples on \dldv~-- } We show novel-view rendering obtained by ours, Gspat-mcmc~\cite{kheradmand20243d}, LVSM~\cite{jin2025lvsm} and Difix3D~\cite{wu2025difix3d+}. Our approach achieves the sharper rendering as well as the better fidelity to the ground-truth.  
}
\label{fig:main-comparison}
\vspace{-1.2em}
\end{figure*}

\subsection{In-domain evaluation}
We evaluate HAD on DL3DV, where we train our hallucination network, thereby providing an in-domain evaluation setting. We compare against feedforward NVS networks (DepthSplat~\cite{xu2025depthsplat}, LVSM~\cite{jin2025lvsm}), two variants of 3DGS, and state-of-the-art diffusion prior-assisted 3DGS pipelines (Difix3D~\cite{wu2025difix3d+} and Difix3D+~\cite{wu2025difix3d+}).
Note Difix3D+, an enhanced version of Difix3D,  further improves image photorealism in the post-rendering stage.

\paragraph{Quantitative results -- \cref{tab:main_result}.} 
Our method outperforms the baselines by a large margin across all metrics. To enable fair comparison with our closest pipeline, we adopt Difix3D's two-phase optimization strategy for 3DGS. In this approach, we first train an initial 3DGS model using only the input views, then optimize the model using both input and augmented views. Under this setting, we replace Difix3D's diffusion prior with our HAD. We demonstrate that even under this suboptimal configuration, our method surpasses the state of the art, highlighting the critical importance of hallucination awareness in diffusion priors.
We further show that our method outperforms Difix3D in the post-rendering setting, where the diffusion prior refines the rendered images of the final 3DGS model. Notably, while the post-rendering improves the photorealism metric (LPIPS), it degrades fidelity metrics (PSNR and SSIM) for both Difix3D and our method.

\paragraph{Qualitative results -- \cref{fig:main-comparison}.}
Qualitatively, we demonstrate that HAD generates augmented views with both photorealism and high consistency with the input views. Notably, LVSM~\cite{jin2025lvsm}, despite producing blurry images, achieves better fidelity, validating our approach of leveraging LVSM's feature backbone as multi-view encoder for hallucination score prediction.

\begin{table}
\centering
\vspace{-0.8em}
\caption{ \textbf{Quantitative comparison of different methods on \textbf{MipNerf360} \cite{barron2022mip}.} Note the results of Genfusion and FSGS are from Genfusion~\cite{wu2025genfusion}.
}
\vspace{-0.5em}
\resizebox{\columnwidth}{!}{
\begin{tabular}{lccccc}
\toprule
\textbf{Method} & \textbf{Type} & \textbf{PSNR}$\uparrow$ & \textbf{SSIM}$\uparrow$ & \textbf{LPIPS}$\downarrow$ \\
\midrule
Gsplat-3DGS \cite{kerbl3Dgaussians} & Optimization & 15.748 & 0.424 & 0.431 \\
Gsplat-mcmc \cite{kheradmand20243d} & Optimization & 17.102 & 0.454 & \cellcolor{3rd} 0.385 \\
FSGS \cite{zhu2024fsgs} & Optimization & 17.940 & 0.492 & 0.468 \\
GenFusion \cite{wu2025genfusion} & Optimization & \cellcolor{2nd}{18.360} & \cellcolor{2nd}{0.496} & {0.465} \\
Difix3D \cite{wu2025difix3d+} & Optimization & \cellcolor{3rd}{18.001} & \cellcolor{3rd}{0.475} & \cellcolor{2nd}{0.350} \\
Ours & Optimization & \cellcolor{1st}{18.689} & \cellcolor{1st}{0.5094} & \cellcolor{1st}{ 0.334} \\
\midrule
Difix3D+ \cite{wu2025difix3d+} & Post-rendering & \cellcolor{2nd}17.571 & \cellcolor{2nd}0.433 & \cellcolor{2nd} 0.326 \\
Ours+ & Post-rendering& \cellcolor{1st}{18.137} & \cellcolor{1st}{0.456} & \cellcolor{1st}{ 0.309} \\
\bottomrule
\end{tabular}
}
\vspace{-1.2em}
\label{tab:main_result_mipnerf360}
\end{table}
\subsection{Cross-domain evaluation}
We also examine whether our HAD can generalize to a different dataset -- \mipnerf. We compare against baselines including FSGS~\cite{zhu2024fsgs}, GenFusion~\cite{wu2025genfusion} and Difix3D~\cite{wu2025difix3d+}. 
As shown in \cref{tab:main_result_mipnerf360}, similar to the in-domain evaluation, our method achieves state-of-the-art performance. 
Note that GenFusion~\cite{wu2025genfusion} leverages video diffusion priors that exhibit better multi-view consistency, thereby outperforming baselines using image diffusion priors such as Difix3D. Nevertheless, our method demonstrates the better performance due to our hallucination-aware strategy. \ws{As shown in \cref{sec:more_diffusion_results}, we further validate that HAD generalizes across different diffusion models (e.g., video and multi-view diffusion~\cite{wu2025genfusion, zhou2025stable}).}

\subsection{Ablation studies}
We conduct ablation studies on different components, the number of versions and fusion strategy in the multi-sampling strategy (M.S.), the pretrained multiview encoder, and the performance of our method under dense view settings. We report the performance on \dldv. Except for the dense view setting where we use 24 views, all ablation studies employ the 9-views setting.

\begin{table}
\centering
\caption{\textbf{Impact of different components}. 
}
\vspace{-0.5em}
\resizebox{0.8\columnwidth}{!}{
\begin{tabular}{lccc}
\toprule
Method & PSNR $\uparrow$ & SSIM $\uparrow$ & LPIPS $\downarrow$ \\
\midrule
Gsplat-mcmc & 20.532 & 0.721 & 0.225 \\
Difix3D & 21.355 & 0.734 & 0.199 \\
Difix3D + HAD & 21.779 & 0.749 & 0.195 \\
Difix3D + HAD + M.S. & 21.983 & 0.755 & 0.195 \\
\textbf{Ours} (full model) & \textbf{22.134} & \textbf{0.757} & \textbf{0.190} \\
\bottomrule
\end{tabular}
}
\label{tab:main-ablation}
\vspace{-0.5em}
\end{table}

\paragraph{Impact of different components -- \cref{tab:main-ablation}.}
As shown in \cref{tab:main-ablation}, HAD notably improves performance over Difix3D, confirming the importance of hallucination awareness. The multi-sampling strategy further enhances reconstruction consistency and perceptual quality. In addition, thanks to the high fidelity of augmented views from HAD, we simplify 3DGS training to a single-phase approach, differing from Difix3D which employs a two-phase strategy.  

\begin{table}
\caption{\textbf{Number of versions in multi-sampling strategy}. }
\vspace{-0.5em}
\centering
\resizebox{0.8\columnwidth}{!}{
\begin{tabular}{lccc}
\toprule
Num. of versions & PSNR $\uparrow$ & SSIM $\uparrow$ & LPIPS $\downarrow$ \\
\midrule
1 (No M.S.) & 21.779 & 0.749 & 0.195 \\
2 & 21.938 & 0.754 & 0.195 \\
3 & \textbf{21.983} & \textbf{0.755} & \textbf{0.195} \\
\bottomrule
\end{tabular}
}
\vspace{-0.5em}
\label{tab:numviews_MS_ablation}
\end{table}

\paragraph{Impact of the number of versions in M.S. --\cref{tab:numviews_MS_ablation}.}
We can create multiple versions of augmented views with diffusion priors by conditioning on different sampled input images and then fusing them into a better final view. As shown in \cref{tab:numviews_MS_ablation}, the more versions we create, the better the performance. We select 3 views to achieve a trade-off between marginal improvement and computational overhead.

\begin{table}
\caption{\textbf{Fusion methods in multi-sampling strategy}. We compare two approaches: (1) \texttt{ArgMin}: selecting pixels with the lowest hallucination score; (2) \texttt{Weighted Average}: computing the weighted mean.}
\vspace{-0.5em}
\centering
\resizebox{0.8\columnwidth}{!}{
\begin{tabular}{lccc}
\toprule
Method & PSNR $\uparrow$ & SSIM $\uparrow$ & LPIPS $\downarrow$ \\
\midrule
 \texttt{Weighted Average}  & 21.856 & 0.747 & 0.190 \\
 \texttt{ArgMin} (Ours) & \textbf{22.134} & \textbf{0.757} & \textbf{0.190} \\
\bottomrule
\end{tabular}
}
\vspace{-0.5em}
\label{tab:abs_fusion_MS}
\end{table}

\paragraph{Fusion strategy in M.S. -- \cref{tab:abs_fusion_MS}.}
We study how to fuse the multi-sampling results. We evaluate two strategies: weighted averaging based on hallucination scores maps and ArgMin fusion, which selects the pixel with the lowest hallucination score across samples. As shown in \cref{tab:abs_fusion_MS}, ArgMin fusion achieves the best overall performance, producing sharper and more consistent reconstructions.

\begin{table}
\centering
\caption{ \textbf{Different hallucination score estimators}. 
We use Mean Absolute Error (MAE) of the predicted hallucination score maps as our evaluation metric. We demonstrate that our hallucination score network, \textit{with the pretrained multiview encoder}, achieves the best performance.
}
\resizebox{0.8\columnwidth}{!}{
\begin{tabular}{lccc}
\toprule
Method & Retrained Difix3D & Ours (w/o pretrained )& Ours (full) \\
\midrule
MAE $\downarrow$ & 0.058 & 0.054 & \textbf{0.043} \\
\bottomrule
\end{tabular}
}
\vspace{-1.2em}
\label{tab:diff_conf}
\end{table}
\paragraph{Different hallucination score estimators -- \cref{tab:diff_conf}.}
We study the performance of three hallucination score estimators: retrained Difix3D, ours without the pretrained multi-view encoder, and our full method. For the retrained Difix3D, we follow Difix3D~\cite{wu2025difix3d+} to finetune the diffusion prior to augment novel views while predicting hallucination scores. We evaluate the performance of hallucination estimation on our curated dataset of 114 training scenes and 26 testing scenes. We calculate the Mean Absolute Error (MAE) of the predicted hallucination score maps on the test scenes. As shown in~\cref{tab:diff_conf}, the finetuned diffusion prior, due to its lack of multi-view reasoning ability, predicts the worst hallucination scores. Similarly, our method without the pretrained multi-view encoder performs worse. This comparison demonstrates the importance of leveraging multi-view reasoning ability of a pretrained NVS network.

\begin{table}
\centering
\caption{\textbf{The performance of our method in dense view setting (24 views).}}
\vspace{-0.5em}
\resizebox{0.8\columnwidth}{!}{
\begin{tabular}{lccc}
\toprule
Method (GSplat-MCMC) & PSNR $\uparrow$ & SSIM $\uparrow$ & LPIPS $\downarrow$ \\
\midrule
Gsplat-MCMC & 26.280 & 0.869 & 0.101 \\
Difix3D & 26.770 & 0.874 & 0.0926 \\
Ours & \textbf{26.969} & \textbf{0.876} & \textbf{0.0921} \\
\bottomrule
\end{tabular}
}
\label{tab:denser-view-dl3dv}
\vspace{-1.5em}
\end{table}

\paragraph{The performance of our method in dense view setting -- \cref{tab:denser-view-dl3dv}.}
We also examine whether our method remains effective when the input coverage becomes denser. On \dldv with 24 input views, the Difix3D already achieves a high PSNR of 26.77 dB, where hallucination artifacts are largely reduced. 
Even in this setting, our method still improves PSNR by 0.20 dB over Difix3D and by 0.69 dB over the original 3DGS model, indicating that the hallucination-aware diffusion prior continues to provide benefits beyond the sparse-view regime.

\section{Conclusion and Future Work}
In this work,  we identify and address a critical limitation in diffusion-assisted 3D reconstruction: while diffusion priors effectively alleviate data sparsity, they introduce hallucinated content that compromises fidelity to input views despite achieving photorealistic rendering. We propose Hallucination-Aware Diffusion priors (HAD), the first framework to explicitly tackle the hallucination issue in diffusion-assisted 3D reconstruction through hallucination score modeling. 
The experimental results on both in-domain and cross-domain benchmarks demonstrate that HAD achieves state-of-the-art performance, with consistent improvements resulting from effective hallucination mitigation. Our work establishes hallucination awareness as a crucial component for diffusion-assisted 3D reconstruction.

\paragraph{Limitations and future works.}   
Our method is bounded by the pretrained multi-view encoder. An interesting direction for future work is to scale up the training of our model by removing the need for complex data requirements--for instance, using uncalibrated multi-view images as in Rayzer~\cite{jiang2025rayzer}. 
\ws{Additionally, we validate our method only on reconstruction of seen regions, and the extension of hallucination-aware modeling to improve generation in unseen regions represents an important direction for future research.}

\newpage
{
    \small
    \bibliographystyle{ieeenat_fullname}
    \bibliography{main}
}


\clearpage
\setcounter{page}{1}
\maketitlesupplementary

\paragraph{Overview.} This supplementary material provides more analysis and experimental validation of our proposed HAD (Hallucination-Aware Diffusion) framework. 
In \cref{sec:hallucination—analysis_appendix}, we analyze hallucination patterns in NVS tasks across two diffusion paradigms -- i.e., diffusion-assisted NVS with explicit 3DGS model and NVS directly via diffusion. 
In \cref{sec:model_arch_appendix}, we detail the architecture, training dataset curation, and implementation of our hallucination scoring network. 
In \cref{sec:more_diffusion_results}, to demonstrate generalization, we show that our scoring network effectively identifies hallucinations in video diffusion (GenFusion~\cite{wu2025genfusion}) and multi-view diffusion (SVC~\cite{zhou2025stable}) without fine-tuning. We further show the quantitative improvement of GenFusion in NVS via diffusion-assisted 3DGS training. 
In \cref{sec:more_results}, we provide additional qualitative comparisons on DL3DV and MipNeRF360 datasets, along with ablation studies validating our design choices, particularly the importance of pretrained initialization for reliable hallucination detection.

\section{Hallucination in NVS via diffusion}
\label{sec:hallucination—analysis_appendix}
\begin{figure*}
\vspace{-0.5em}
  \centering
  \includegraphics[width=\linewidth]{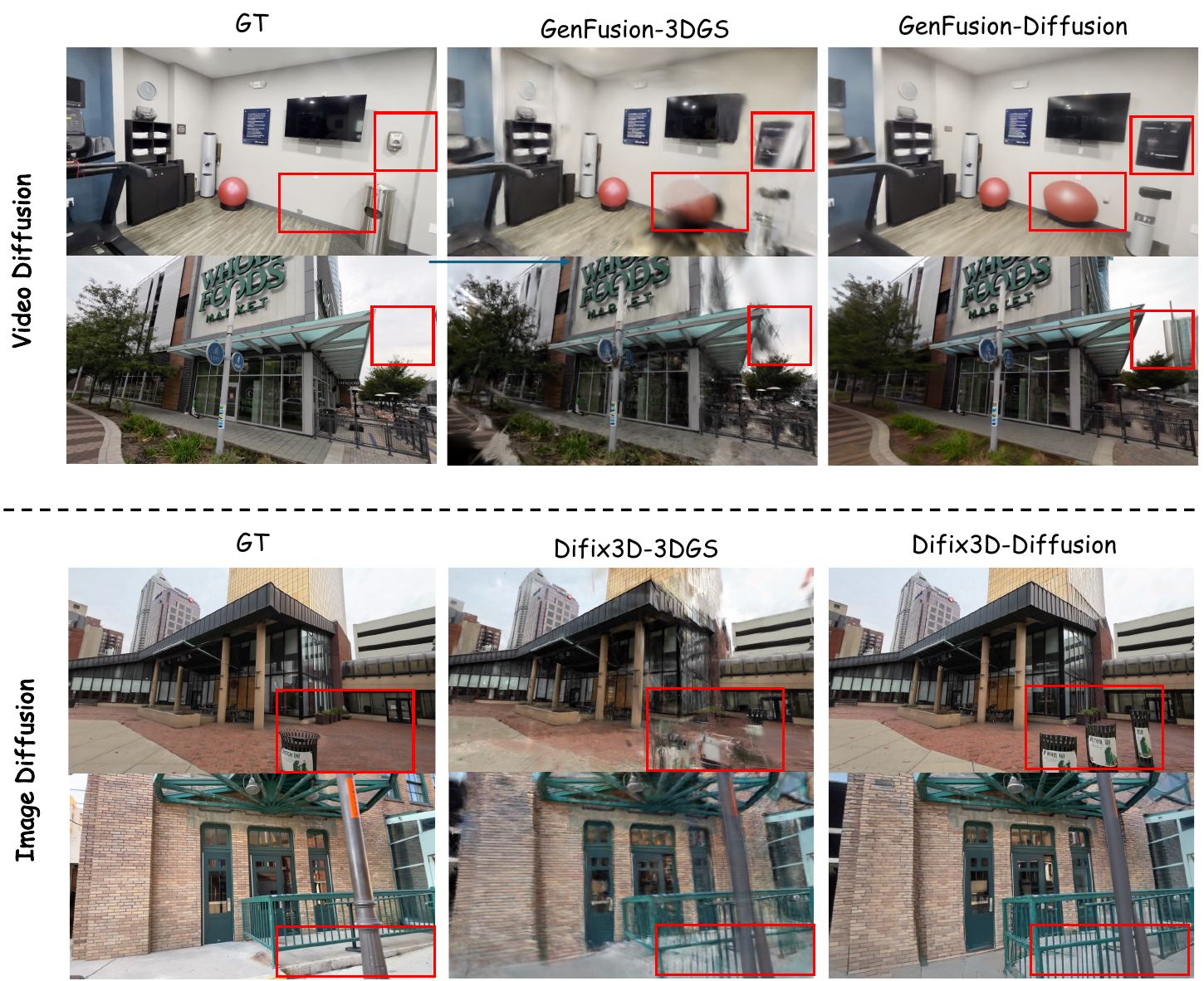}
\caption{Qualitative demonstration of hallucination pattens in diffusion-assisted 3DGS pipelines. Both Difix3D (image diffusion) and GenFusion (video diffusion) iteratively enhance the 3DGS rendering. As a result, even mild artifacts in the initial 3DGS output, such as small floaters or subtle geometric distortions, can be progressively amplified by diffusion priors and eventually evolve into clearly visible hallucinations. This demonstrates that hallucination is not introduced abruptly, but can accumulate and become more pronounced during iterative refinement. \textbf{Note:} \textit{method-3DGS} denotes direct rendering from 3DGS, while \textit{method-diffusion} denotes diffusion-enhanced results.}
\label{fig:diffusion_3dgs}
\vspace{-1.2em}
\end{figure*}

\begin{figure*}
\vspace{-0.5em}
  \centering
  \includegraphics[width=\linewidth]{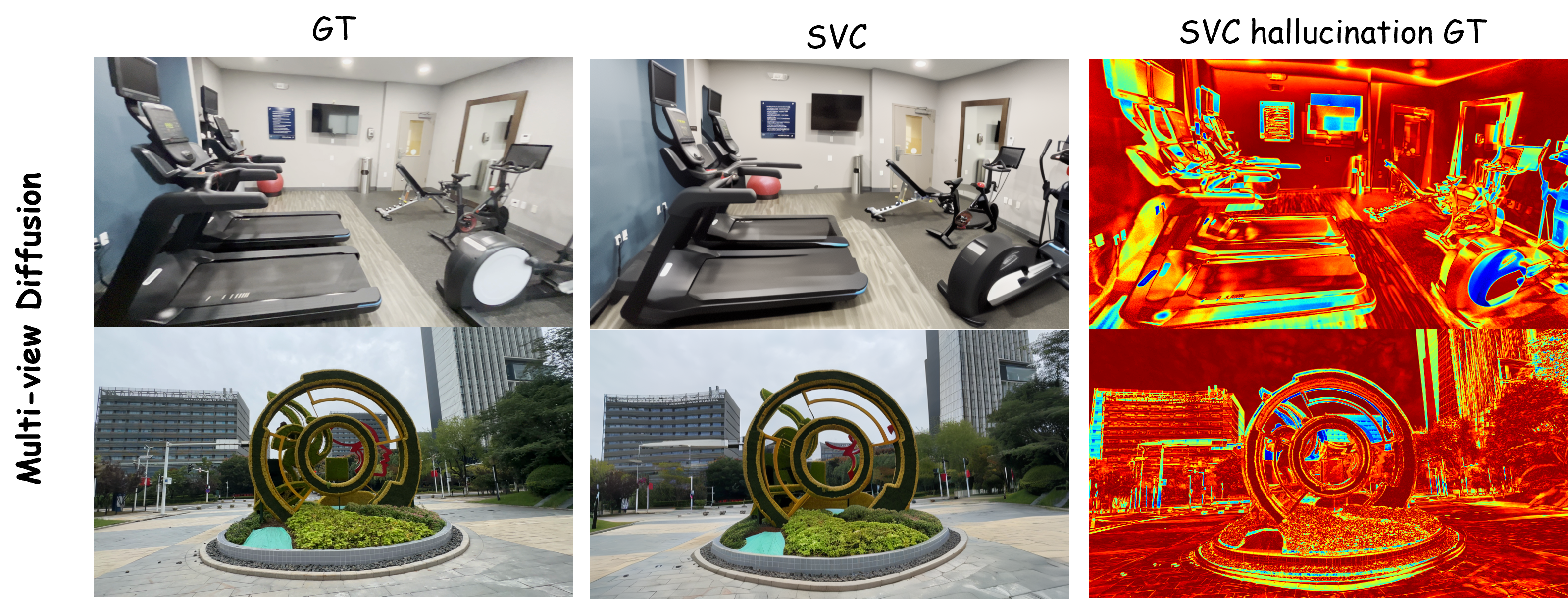}
    \caption{Hallucination analysis for multi-view diffusion (SVC). Hallucination maps are computed against ground-truth images to highlight inconsistencies. We observe that hallucinations mainly arise from inaccurate 3D structure, where objects exhibit misaligned positions and distorted geometry across views due to the absence of explicit 3D constraints.}
\label{fig:mv-diffusion}
\vspace{-0.5em}
\end{figure*}
We provide additional analysis to deepen understanding the hallucination issue introduced by diffusion models in NVS task. 
Specifically, we evaluate recent state-of-the-art methods from two representative paradigms: \textit{Diffusion-assisted NVS with explicit 3DGS model} (e.g., Difix3D~\cite{wu2025difix3d+}, GenFusion~\cite{wu2025genfusion} and 3DGS-enhancer~\cite{liu20243dgs}), and \textit{NVS directly via diffusion} such as SVC~\cite{zhou2025stable}. We consistently observe hallucinations across both paradigms, regardless of whether image diffusion, video diffusion, or direct diffusion-based synthesis is employed.

\paragraph{Diffusion-assisted NVS with explicit 3DGS model.} As shown in \cref{fig:diffusion_3dgs}, hallucinations mainly arise from imperfect geometry in the underlying 3D representation. Specifically, in regions with sparse observations or unseen viewpoints, 3DGS often produces floaters or severely distorted structures. When applied to such renderings, diffusion models tend to ``correct" these artifacts by introducing semantically plausible but incorrect content, often borrowing from reference views, thereby amplifying inconsistencies and causing hallucinated geometry and appearance.

\paragraph{NVS directly via diffusion.} As shown in ~\cref{fig:mv-diffusion}, SVC~\cite{zhou2025stable}, while achieving the photorealistic rendering, still suffers from a different mode of hallucination. As they directly rely on diffusion models without explicit 3D representations, the lack of geometric constraints often leads to structural distortions and inconsistent geometry across views, leading to the geometrically inconsistent structures across viewpoints.

\section{Details of Hallucination Scoring Network}
\begin{figure}
\vspace{-0.5em}
  \centering
  \includegraphics[width=\linewidth]{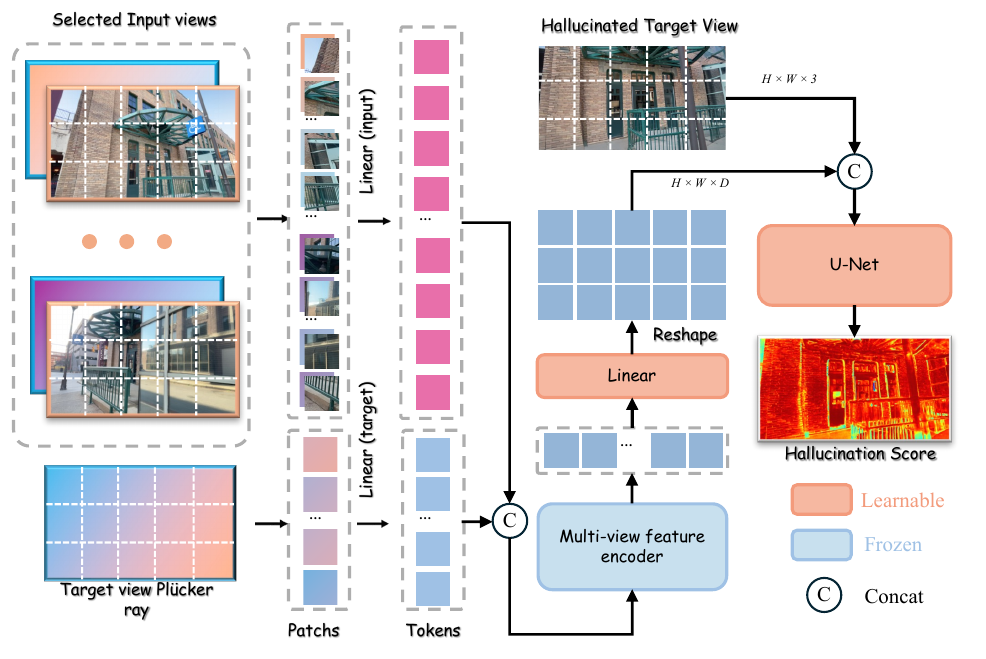}
\caption{
{\bf Overview of hallucination scoring network -- } 
The network predicts a pixel-wise hallucination score map $\hallucinationscore$ for a hallucinated novel view $\tilde{\image}_{\diffusion}$. 
It consists of a multi-view feature encoder $\multiviewencoder$ (the frozen feature backbone of a pre-trained LVSM) 
and a three-layer U-Net score branch $\scoring$, which estimates hallucination scores using both multi-view 
features and the novel view image. The model is trained on curated multi-view and hallucinated novel-view pairs.
}
\label{fig:hallucination_network}
\vspace{-1.2em}
\end{figure}
\label{sec:model_arch_appendix}

\paragraph{Overview of hallucination score network.} We provide a detailed model architecture \cref{fig:hallucination_network}.

\paragraph{Training dataset curation.}
We provide additional details on the constructing training dataset for the hallucination score network. For all training scenes, we follow the Difix3D~\cite{wu2025difix3d+} pipeline under the 9-view setting to first reconstruct a 3DGS model from sparse input views. 
We then render all remaining views that are not included in the 9 input views, obtaining the corresponding 3DGS renderings. 
For each such view, we further apply the diffusion-based refinement module in Difix3D to generate enhanced images. This process results in a triplet of aligned images $(I_{\text{GT}}, I_{\text{difix}}, I_{\text{3DGS}})$ for each viewpoint:
\begin{itemize}
    \item \textbf{$I_{\text{GT}}$}: the ground-truth image from the captured view.
    \item \textbf{$I_{\text{3DGS}}$}: the rendering from the reconstructed 3DGS model, which often contains artifacts such as floaters or distorted structures due to incomplete geometry. 
    \item \textbf{$I_{\text{difix}}$}: the diffusion-enhanced result, where diffusion models attempt to refine the 3DGS rendering but may introduce hallucinated content.
\end{itemize}

These triplets $(I_{\text{GT}}, I_{\text{difix}}, I_{\text{3DGS}})$ and corresponding camper pose $\pose$ serve as supervision for training the hallucination score network, enabling it to learn to identify hallucination introduced by diffusion model.

\paragraph{Training details.}
Given the constructed triplets, we train the hallucination score network based on the pretrained LVSM encoder. For each triplet, we use its camera pose as the target view, and select the three nearest input views from the training views used in Difix3D+ (i.e., the 9 input views for 3DGS reconstruction) based on camera pose proximity. The selected input views are encoded using the multiview encoder to extract multi-view features $\multiviewfeature$. The diffusion-enhanced image ($I_{\text{difix}}$) of the triplet is treated as the hallucinated target view. We concatenate the multi-view features with the target-view features extracted from $I_{\text{difix}}$, and feed them into a learnable U-Net to predict the hallucination score map. Optionally, $I_{\text{3DGS}}$ can also be included as an additional input. The network is trained to estimate the MAE between the hallucinated view and the corresponding ground-truth image ($I_{\text{GT}}$), enabling it to identify hallucination regions introduced by diffusion models.

\section{Generalizing to other diffusion models}
\label{sec:more_diffusion_results}
\begin{figure*}
\vspace{-0.5em}
  \centering
  \includegraphics[width=\linewidth]{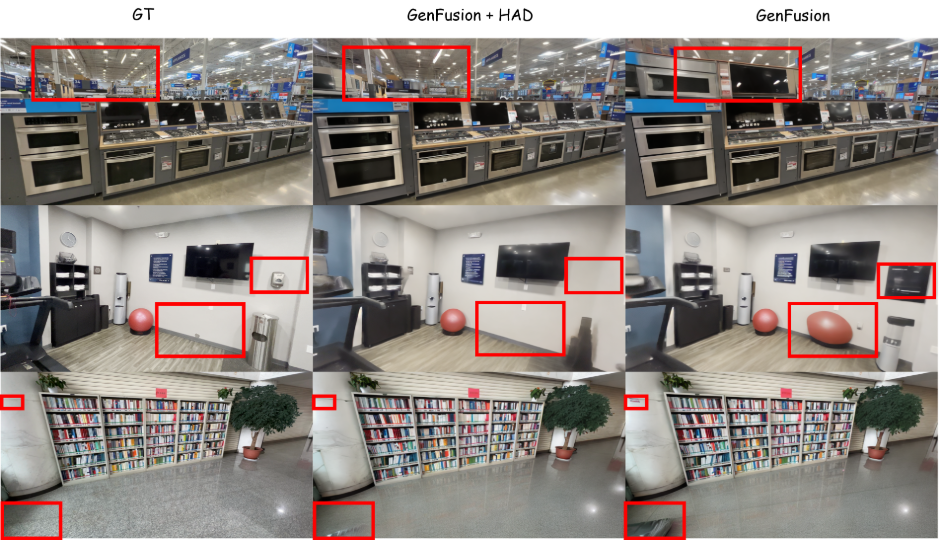}
\caption{
\textbf{Hallucination Scoring for GenFusion.} Our hallucination scoring network can also mitigate hallucinations in video diffusion without fine-tuning.
}
\label{fig:Genfusion-HAD}
\vspace{-0.5em}
\end{figure*}

\begin{table}
\centering
\caption{\textbf{Improving GenFusion~\cite{wu2025genfusion} via HAD}. We demonstrate that our hallucination scoring network generalizes to video diffusion models without fine-tuning, effectively masking hallucinated pixels and improving reconstruction quality. We test it on DL3DV~\cite{ling2024dl3dv}}
\vspace{-0.5em}
\resizebox{0.8\columnwidth}{!}{
\begin{tabular}{lccc}
\toprule
Method & PSNR $\uparrow$ & SSIM $\uparrow$ & LPIPS $\downarrow$ \\
\midrule
Genfusion & 20.57 & 0.7396 & 0.2845 \\
Genfusion + HAD & \textbf{20.80} & \textbf{0.7415} & \textbf{0.2817} \\
\bottomrule
\end{tabular}
}
\label{tab:abs-genfusion}
\vspace{-1.5em}
\end{table}

\subsection{Improving GenFusion}
We integrate our hallucination scoring network into GenFusion~\cite{wu2025genfusion} -- the state-of-the-art video-diffusion-assisted 3DGS training pipeline. Importantly, we apply the same HAD model as in the main paper without any additional fine-tuning on video diffusion data. To ensure a fair comparison, we keep all original GenFusion settings and only incorporate our hallucination scoring module, testing on the DL3DV dataset~\cite{ling2024dl3dv}.  
Specifically, for each generated novel view, HAD predicts a hallucination score map and mask out pixels with low confidence, following the same strategy as Difix3D + HAD. As shown in \cref{tab:abs-genfusion}, this simple integration leads to a PSNR improvement of +0.23.  The qualitative results in \cref{fig:Genfusion-HAD} shows that our method effectively reduces hallucinated content and improves reconstruction fidelity.

\subsection{Hallucination Detection in SVC and GenFusion}

\begin{figure*}
 \vspace{-0.5em}
  \centering
  \includegraphics[width=\linewidth]{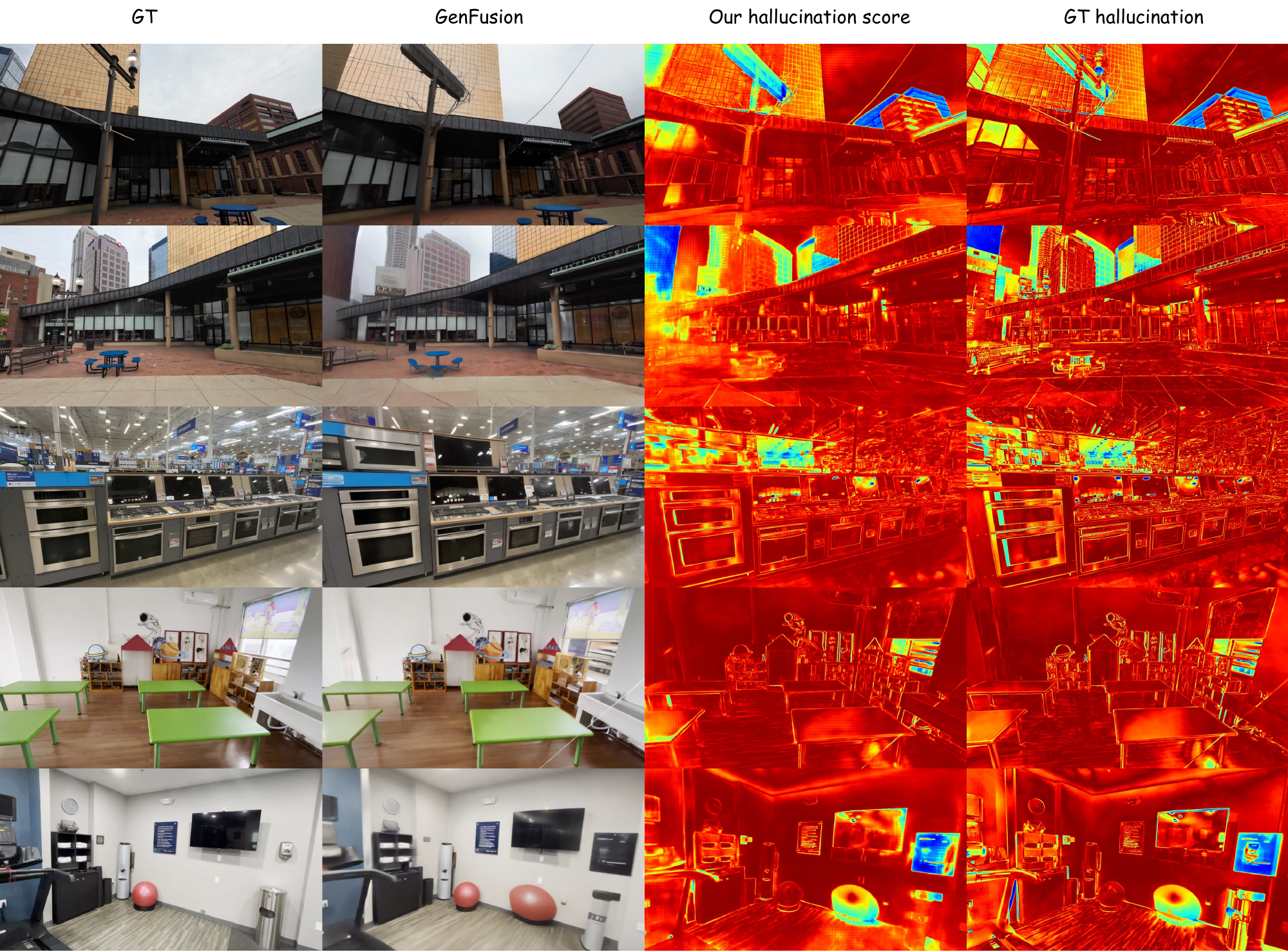}
    \caption{Generalization of our hallucination scoring network to video diffusion (GenFusion). Our model is \textbf{not fine-tuned} on the target diffusion model, demonstrating strong generalization across diffusion paradigms.}
    \vspace{-0.5em}
\label{fig:genfusion}
\end{figure*}

\begin{figure*}
 \vspace{-0.5em}
  \centering
  \includegraphics[width=\linewidth]{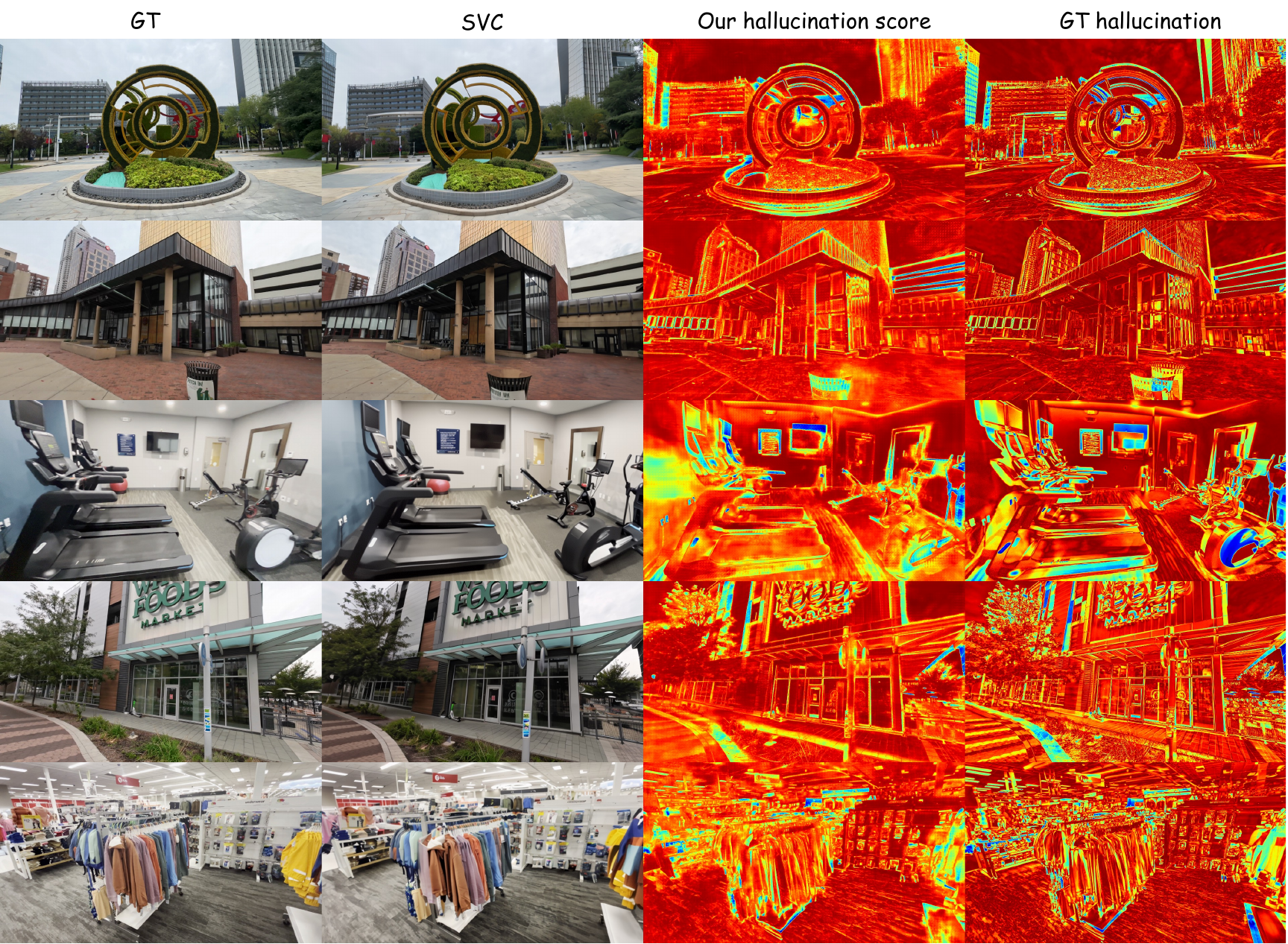}
    \caption{Generalization of our hallucination scoring network to multi-view diffusion (SVC). Our model is \textbf{not fine-tuned} on the target diffusion model, demonstrating strong generalization across diffusion paradigms.}
    \vspace{-0.5em}
\label{fig:SVC}
\end{figure*}
To demonstrate the generalization beyond Difix3D, we apply the hallucination scoring network other diffusion-based pipelines, including video diffusion (GenFusion~\cite{zhou2025stable}), multi-view diffuison (SVC~\cite{zhou2025stable}). As shown in ~\cref{fig:genfusion} and ~\cref{fig:SVC}, our hallucination scoring network is able to effectively identify hallucination regions introduced by different diffusion models, despite not being trained on these data. This demonstrates the strong generalization capability across diverse diffusion paradigms.

\section{More Results}
\label{sec:more_results}
\subsection{Additional Qualitative Comparisons}
\begin{figure*}
\vspace{-0.5em}
  \centering
  \includegraphics[width=0.95\linewidth]{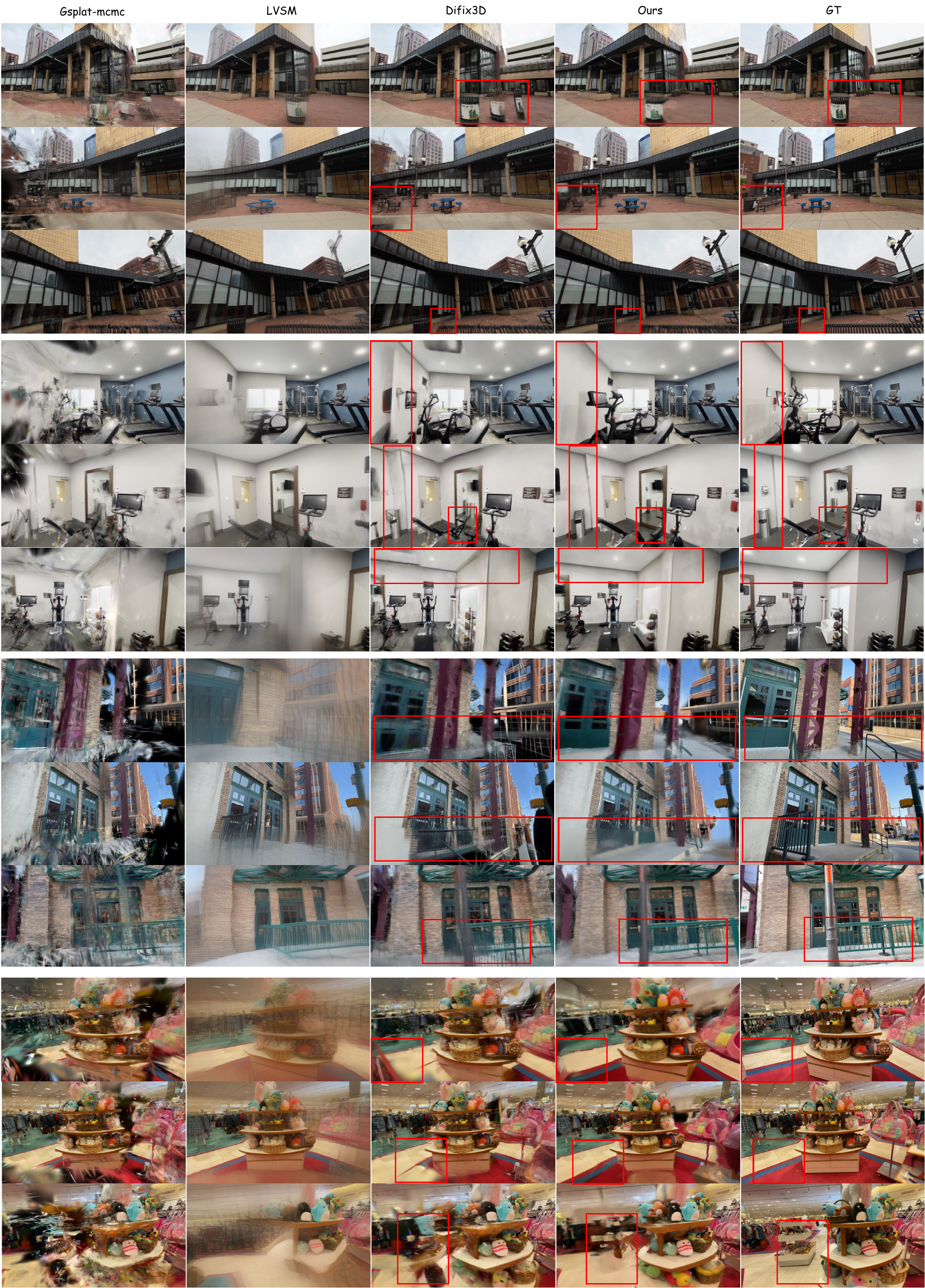}
\caption{
\textbf{More Qualitative Results on DL3DV \cite{ling2024dl3dv}.}
}
\label{fig:DL3DV_v2}
\vspace{-1.2em}
\end{figure*}

\begin{figure*}
\vspace{-0.5em}
  \centering
  \includegraphics[width=\linewidth]{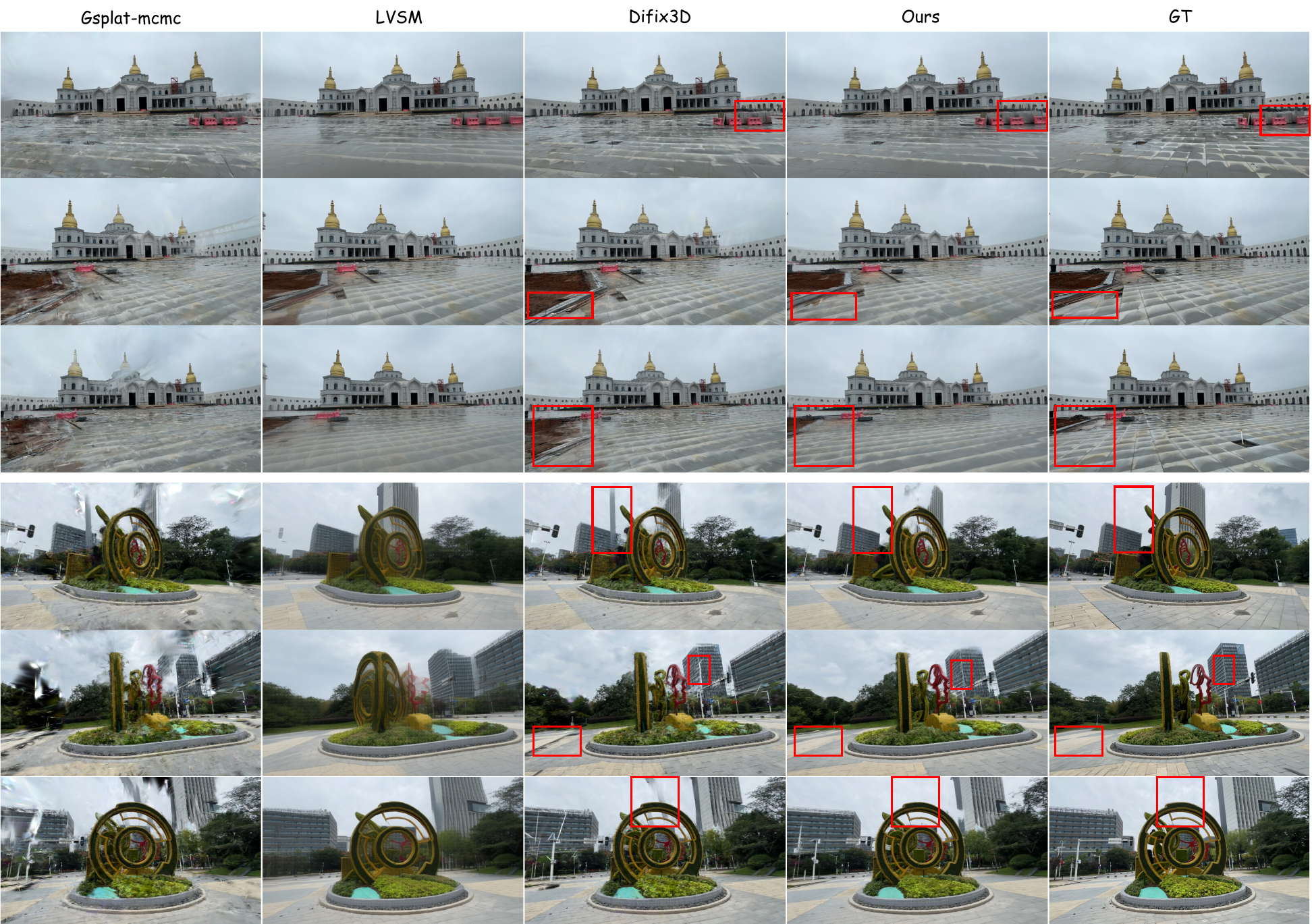}
\caption{
\textbf{More Qualitative Results on DL3DV \cite{ling2024dl3dv}.}
}
\label{fig:DL3DV_v1}
\vspace{-1.2em}
\end{figure*}

\begin{figure*}
\vspace{-0.5em}
  \centering
  \includegraphics[width=\linewidth]{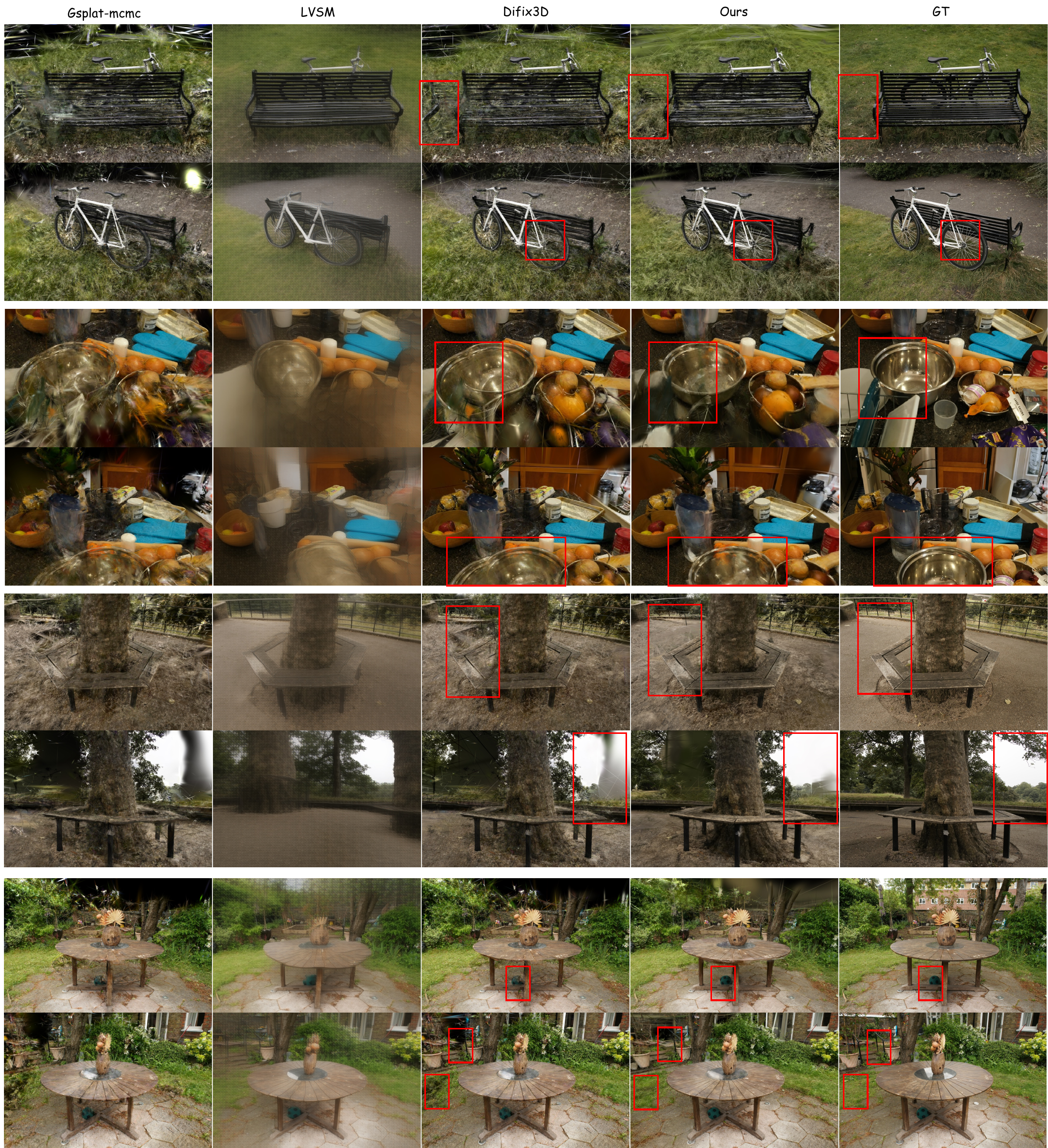}
\caption{
\textbf{More Qualitative Results on MipNeRF-360 \cite{barron2022mip}.}
}
\label{fig:MipNeRF_360}
\vspace{-1.2em}
\end{figure*}

We provide additional qualitative results on both the DL3DV -- as shown in \cref{fig:DL3DV_v1} and \cref{fig:DL3DV_v2}, and MipNeRF360 datasets -- see \cref{fig:MipNeRF_360}. 
Note we also include the corresponding rendered videos in project website, providing a clearer comparison across viewpoints.
Both the qualitative results and videos show that our method produces more geometrically consistent renderings with fewer hallucinated structures compared to the baselines.

\subsection{More Ablation Studies}

We study how two factors affect the performance of our scoring network: using pre-trained weights for the multi-view encoder $\multiviewencoder$ and adding the 3DGS-rendered image $\rendering_{\gsparam}$ as an extra input. To evaluate the effect of pre-training, we train the multi-view encoder from scratch on the same dataset. To examine the role of the 3DGS-rendered image, we remove this input and only use the multi-view features $\multiviewfeature$ and diffusion-generated images $\tilde{\image}_{\diffusion}$.

\begin{table}
\centering
\caption{ 
We compare different variants of hallucination score network including our model without 3DGS rendering input, our model without pretrained weights, and our full model in score estimation accuracy.
}
\resizebox{0.8\columnwidth}{!}{
\begin{tabular}{lccc}
\toprule
Method & Ours (w/o 3dgs input) & Ours (w/o pretrained )& Ours (full) \\
\midrule
MAE $\downarrow$ & 0.044 & 0.054 & \textbf{0.043} \\
\bottomrule
\end{tabular}
}
\vspace{-1em}
\label{tab:abl_hal-net}
\end{table}
\begin{table}
\centering
\caption{The impact of different design choices in hallucination score network on the 3D reconstruction.}
\resizebox{0.8\columnwidth}{!}{
\begin{tabular}{lccc}
\toprule
Method & PSNR $\uparrow$ & SSIM $\uparrow$ & LPIPS $\downarrow$ \\
\midrule
Ours & \textbf{22.134} & \textbf{0.757} & \textbf{0.190} \\
W/o pretrain& 21.600 & 0.748 & 0.1974 \\
W/o input of 3dgs & 21.960  & 0.755 & 0.1891 \\
\bottomrule
\end{tabular}
}
\label{tab:abs_lvsm}
\vspace{-1.2em}
\end{table}
As shown in ~\cref{tab:abs_lvsm} and \cref{tab:abl_hal-net} where we  study the impact on score estimation accuracy and the final 3D reconstruction, removing the pre-trained initialization leads to a clear performance drop, showing that the strong 3D-awareness is essential for predicting reliable hallucination scores. In contrast, excluding the 3DGS-rendered input results in only a minor change, indicating that this cue is helpful but not crucial for our scoring network.

\end{document}